\documentclass[letterpaper, 10 pt, conference]{ieeeconf}  

\IEEEoverridecommandlockouts                              
\usepackage{graphicx}
\usepackage{mathtools}
\usepackage{amssymb}
\usepackage{algorithm, algorithmicx, algpseudocode}
\usepackage{epstopdf}
\usepackage{xcolor}
\usepackage{hyperref}
\usepackage{adjustbox} 
\usepackage{caption} 
\usepackage{floatpag}
\usepackage{subcaption}
\usepackage{graphicx}
\usepackage{multirow}
\floatpagestyle{empty}  

\title{\LARGE \bf
AttenGluco: Multimodal Transformer-Based Blood Glucose Forecasting on AI-READI Dataset
}

\author{Ebrahim Farahmand$^{\star}$, Reza Rahimi Azghan$^{\star}$,  Nooshin Taheri Chatrudi$^{\star}$, Eric Kim$^{\star}$, Gautham Krishna Gudur$^{\dagger}$, \\ Edison Thomaz$^{\dagger}$, Giulia Pedrielli$^{\star}$, Pavan Turaga$^{\star}$, Hassan Ghasemzadeh$^{\star}$
\thanks{$^{\star}$Arizona State University, Phoenix, AZ, USA,\vfill$^{\dagger}$ The University of Texas at Austin, Austin, TX, USA}%
}

\begin{document}

\maketitle
\thispagestyle{empty}
\pagestyle{empty}

\begin{abstract}
Diabetes is a chronic metabolic disorder characterized by persistently high blood glucose levels (BGLs), leading to severe complications such as cardiovascular disease, neuropathy, and retinopathy. Predicting BGLs enables patients to maintain glucose levels within a safe range and allows caregivers to take proactive measures through lifestyle modifications. Continuous Glucose Monitoring (CGM) systems provide real-time tracking, offering a valuable tool for monitoring BGLs. However, accurately forecasting BGLs remains challenging due to fluctuations due to physical activity, diet, and other factors. Recent deep learning models show promise in improving BGL prediction. Nonetheless, forecasting BGLs accurately from multimodal, irregularly sampled data over long prediction horizons remains a challenging research problem. In this paper, we propose \textbf{AttenGluco}\footnote{Code base is available at: \href{https://github.com/rzarhmi/AttenGluco}{\textcolor{blue}{https://github.com/rzarhmi/AttenGluco}}}, a multimodal Transformer-based framework for long-term blood glucose prediction. AttenGluco employs cross-attention to effectively integrate CGM and activity data, addressing challenges in fusing data with different sampling rates. Moreover, it employs multi-scale attention to capture long-term dependencies in temporal data, enhancing forecasting accuracy. To evaluate the performance of AttenGluco, we conduct forecasting experiments on the recently released AIREADI dataset, analyzing its predictive accuracy across different subject cohorts including healthy individuals, people with prediabetes, and those with type 2 diabetes. Furthermore, we investigate its performance improvements and forgetting behavior as new cohorts are introduced. Our evaluations show that AttenGluco improves all error metrics, such as root mean square error (RMSE), mean absolute error (MAE), and correlation, compared to the multimodal LSTM model, which is widely used in state-of-the-art blood glucose prediction. AttenGluco outperforms this baseline model by about 10\% and 15\% in terms of RMSE and MAE, respectively.

\end{abstract}
\section{Introduction}
According to the World Health Organization~\cite{abdul2020epidemiology}, the prevalence of type 2 diabetes has increased significantly over the last decades. In 2022, 14\% of adults aged 18 years and older were living with diabetes, double the 7\% reported in 1990~\cite{WHO2024Diabetes}. This increase is attributed to various factors such as sedentary lifestyles, stress, poor diet, and an aging population~\cite{raffin2023sedentary}. As a result, type 2 diabetes poses a significant public health challenge that requires urgent attention and intervention. Poor management of type 2 diabetes can lead to the progression of chronic health complications and an increased risk of both hyperglycemic and hypoglycemic events. Effectively managing blood glucose levels through consistent monitoring and accurate forecasting is crucial as early intervention measures to prevent hyperglycemic and hypoglycemic events. Accurate glucose prediction is essential for optimizing insulin dosages, meal planning, and exercise habits to maintain blood glucose levels within a safe range.\looseness=-1



CGM devices have been developed as an advanced technology to support diabetes management. CGM devices provide valuable insights into blood glucose fluctuations by collecting continuous glucose signals. The CGM data allows patients to monitor fluctuations and trends in their blood glucose levels more effectively by providing real-time blood glucose level measurements. Thus, CGM devices have grown significantly in recent years, making them a widely adopted tool for diabetes prevention. Furthermore, physiological and behavioral variables, such as physical activity levels (e.g., walking or running) and stress levels, affect blood glucose fluctuation~\cite{shuvo2023deep}. Therefore, the accurate forecasting of blood glucose levels can be evaluated by combining BGL signals with other physiological and behavioral variables. This data integration enables a more comprehensive and personalized approach to managing diabetes, especially for individuals with type 2 diabetes.

Recently, artificial intelligence (AI) and machine learning algorithms have played a critical role in the control and prediction of blood glucose levels. These advanced technologies leverage data from CGM devices and integrate with physiological signals, such as stress levels, heart rate, and physical activity signals. By analyzing these complex datasets, the algorithms can identify trends and patterns in blood glucose fluctuations with high accuracy.\looseness=-1


Sequential machine learning models, notably Long Short-Term Memory (LSTM) networks and Gated Recurrent Units (GRUs)~\cite{chung2014empirical}, are extensively employed in forecasting time-series signals due to their ability to capture temporal dependencies. GRU-based models outperform traditional methods in univariate time-series classification tasks\cite{Elsayed_2019}. Moreover, these models are highly effective for time-series forecasting when optimized with suitable algorithms~\cite{makinde2024optimizingtimeseriesforecasting}. These models have also been extensively applied in predicting Type 1 diabetes outcomes~\cite{patil2024modeling}. However, they often struggle to capture long-term dependencies inherent in time-series data, which in turn limits their effectiveness in long-term forecasting~\cite{kim2024comprehensive}. Research indicates that while LSTMs are designed to manage longer sequential correlations compared to traditional RNNs, they still encounter challenges in memorizing extended sequences~\cite{kong2024unlocking}.\looseness=-1

Recently, transformers have emerged as a powerful model for capturing long-term dependencies in time-series data, primarily through the use of attention mechanisms. Unlike traditional models, which have limited memory and struggle with long-term dependencies, transformers leverage attention mechanisms to effectively capture these dependencies~\cite{kong2024unlocking}.  The attention mechanism within transformers allows the model to weigh the importance of different time steps, which enables them to focus on relevant parts of the sequence when making predictions~\cite{vaswani2017attention}. This capability is particularly beneficial in applications such as forecasting BGL in type 2 diabetes, where understanding long-range temporal relationships is crucial. Moreover, transformers are well-suited for handling time-series data collected at varying sampling rates. Traditional models often face challenges when dealing with such data due to inconsistencies in temporal resolution. Transformers, however, can manage these variations effectively~\cite{chen2024pathformer, zhang2024multi}.\looseness=-1

\textbf{Key Limitations and Associated Challenges:} We highlight the key limitations of state-of-the-art work and present the associated challenges of blood glucose prediction here.
\begin{itemize}
    \item Difficulty in achieving accurate long-term blood glucose forecasting.\looseness=-1
    \item Mismatched temporal resolutions in data sources (e.g., CGM readings, physiological, and behavioral variables).\looseness=-1
    \item Limited clinical datasets, especially for populations with type 2 diabetes.

\end{itemize}

To address these challenges, \textit{a novel accurate forecasting algorithm for long-term prediction is required for individuals with type 2 diabetes.} In this paper, an accurate forecasting model using a Transformer architecture is proposed. The Transformer model is made up of a combination of two attention mechanisms (e.g., cross-attention and multi-scale attention). Cross-attention captures long-term dependencies in temporal data and handles the various temporal resolutions in data sources.  The multi-scale attention captures the influence of external time series variables (e.g., physiological signals, behavioral data) on blood glucose levels. Furthermore, to the best of our knowledge, our work is the first to investigate the problem of blood glucose forecasting on AI-READI dataset~\cite{AIREADI2024,AIREADI20242}. The following list summarizes the novel contributions of our work.

\begin{itemize}
    \item \textbf{We propose a Transformer-based architecture with the new attention layers} to forecast blood glucose levels accurately, especially for long-term forecasting.
    \item \textbf{We developed a hybrid attention mechanism of cross-attention and multi-scale attention} to forecast blood glucose levels.
    \item \textbf{We used various body variables}, such as activity along with BGL to enhance BGL prediction precision.
    \item \textbf{We applied our proposed forecasting blood glucose model on the Flagship AI-READI dataset} for patients with type 2 diabetes.
    \item \textbf{We performed multiple experiments} to evaluate the model’s accuracy across different subject cohorts and analyze both its performance gains and forgetting behavior as new cohorts were introduced.

\end{itemize}


    
    

\section{Related Work}

Recent years have seen a surge in blood glucose management technologies. CGM systems, wearable health monitoring devices, and automated insulin delivery systems (AIDS) collectively provide real-time data and partial automation for diabetes care~\cite{WHO2024Diabetes, pasquel2021management}. CGMs offer continuous monitoring of glucose levels, synchronizing with mobile applications for timely alerts on hyperglycemia and hypoglycemia \cite{abdul2020epidemiology}. Wearable sensors further extend coverage to physiological and behavioral metrics, such as heart rate variability and physical activity levels~\cite{raffin2023sedentary, guan2023artificial}. By combining CGM outputs with additional signals, AIDS can regulate insulin dosage more precisely~\cite{lovic2020growing}. However, limitations persist in terms of sensor calibration, missing data, and user non-adherence~\cite{pasquel2021management}.

Early prediction efforts relied on statistical and time-series models, notably Autoregressive Integrated Moving Average (ARIMA)~\cite{xie2020benchmarking}. Although ARIMA and similar approaches are straightforward, they often fail to capture the complex, nonlinear patterns of glycemic fluctuation. Machine learning (ML) techniques, such as support vector regression and random forests, typically reduce forecasting error by 5--15\% compared to ARIMA~\cite{xie2020benchmarking, mujahid2021machine}. However, they still struggle with deeper temporal dependencies over longer prediction windows~\cite{mujahid2021machine}.

Deep learning techniques, widely applied across various domains such as healthcare~\cite{azghan2023personalized,mamun2022multimodal,10857973}, classification tasks~\cite{10780662}, offer improved forecasting accuracy by effectively modeling intricate temporal dependencies and nonlinear patterns in time series data. Furthermore, integrating causal knowledge into learning frameworks~\cite{corazza2023expediting,10644549} can enhance adaptability and facilitate knowledge transfer across different environments. LSTM architectures are proposed to mitigate vanishing and exploding gradients in recurrent neural networks~\cite{ shuvo2023deep}. By gating internal states, LSTMs retain long-term context for extended horizons, outperforming classical ML methods in certain datasets~\cite{shuvo2023deep}. Despite these improvements, LSTM-based models often demand significant computational resources and meticulous tuning, making them less flexible for large-scale or highly variable glucose data~\cite{kong2024unlocking}.
GRUs streamline the gating structure of LSTMs, converging 15--25\% faster for some time-series tasks~\cite{chung2014empirical, Elsayed_2019}. Nevertheless, GRUs still encounter challenges related to sensor inaccuracies, incomplete user logs, and irregular sampling rates \cite{kong2024unlocking}. Hybrid methods that integrate convolutional layers with recurrent modules reduce some errors by 2--4\% \cite{Elsayed_2019}, yet extensive clinical validation for blood glucose forecasting remains limited.

Transformers adopt self-attention instead of recurrent loops which facilitates parallel learning over extensive sequences~\cite{vaswani2017attention}. This approach outperforms RNNs by 5--10\% in mean squared error (MSE) for long-horizon predictions~\cite{kim2024comprehensive, farahmand2024hybrid}. However, many existing implementations assume large, consistent datasets with minimal missing points. Glucose monitoring, conversely, often faces sensor dropouts and user non-adherence, limiting straightforward application~\cite{acuna2023analyzing}. Gluformer~\cite{sergazinov2023gluformer} developed a transformer-driven blood glucose forecasting model by providing uncertainty intervals rather than single-point estimates. Although a 1--2 mg/dL improvement in short-horizon RMSE has been observed, the absence of multi-scale/cross-attention hinders the model’s ability to integrate additional clinical or activity data~\cite{sergazinov2023gluformer}.

In summary, current blood glucose prediction models have notable limitations. ARIMA struggles with nonlinearities \cite{xie2020benchmarking, mujahid2021machine}, while ML models such as support vector regression improve RMSE but fail in long-range forecasting~\cite{xie2020benchmarking, mujahid2021machine}. LSTMs and GRUs improve but suffer from irregular sample rates of various sensors~\cite{kong2024unlocking, shuvo2023deep}. Recurrent models still encounter inefficiencies for long-horizons forecasting~\cite{kong2024unlocking}. Transformers, including Gluformer, introduce multi-head attention but lack effective cross-attention for integrating multimodal data~\cite{zhu2023edge, acuna2023analyzing}. A more robust approach combining multi-scale and cross-attention is needed for accurate, real-world glucose forecasting.

\section{Proposed Method}
\begin{figure*}[t]
    \centering    \includegraphics[width=0.8\textwidth]{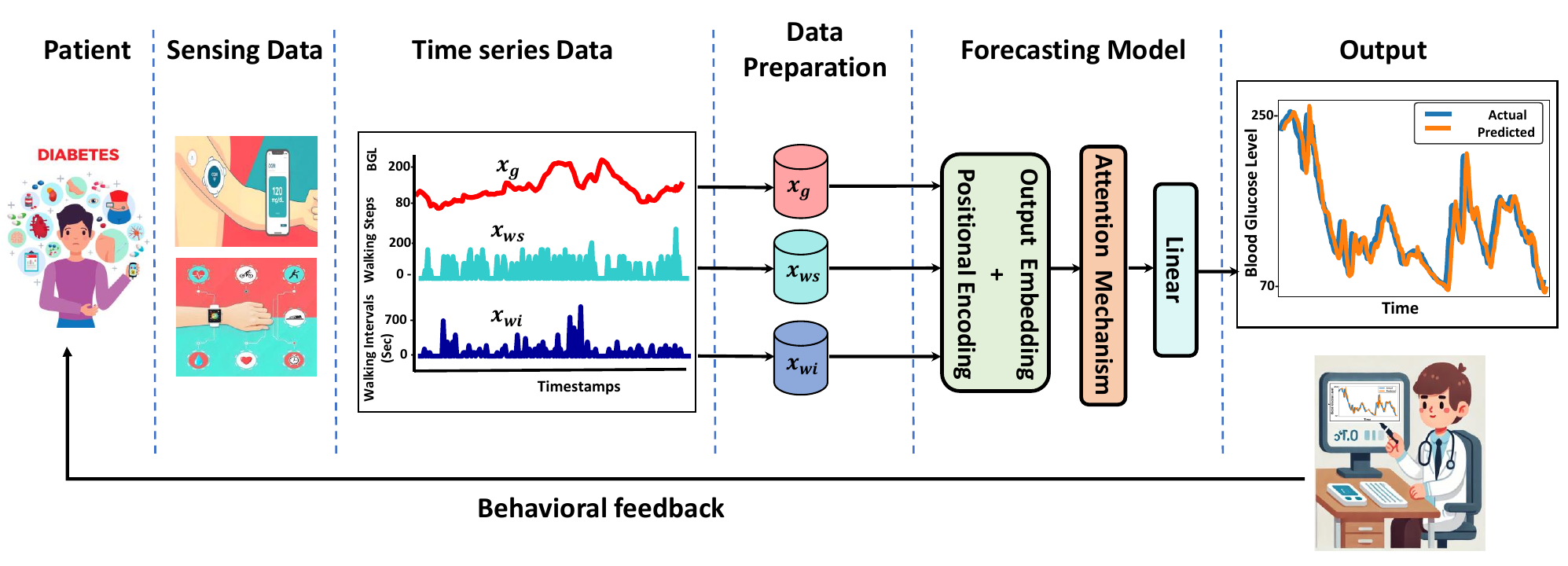}
    \caption{\small Overview of the AttenGluco framework including sensing module, data preparation, and forecasting model.}
    \label{fig:proposed_method}
\end{figure*}

In this section, we introduce our proposed framework for blood glucose prediction. An overview of the AttenGluco framework is shown in Fig.~\ref{fig:proposed_method}. The framework comprises three main components: (a) a sensing module that gathers physiological and behavioral signals from wearable sensors, (b) a preprocessing module for time-series data preparation, and (c) a machine learning forecasting model utilizing the Transformer architecture for blood glucose prediction. Our transformer-based model predicts blood glucose levels (BGL) in individuals with type 2 diabetes by incorporating CGM data alongside activity information.
The attention mechanism within the Transformer facilitates the effective integration of multi-time series signals recorded at different sampling rates. Additionally, it is well-suited for predicting highly fluctuating signals such as BGLs. To validate the effectiveness of our proposed model, we conduct experiments using the publicly available AI-READI (Flagship) dataset. The following sections provide a detailed explanation of the forecasting problem and key components of AttenGluco.

\subsection{Forecasting Problem}
The problem of blood glucose forecasting with multimodal input data can be formulated as a time series prediction task. Let \( \mathbf{X} = [\mathbf{x}_1, \mathbf{x}_2, \dots, \mathbf{x}_k] \) represent a set of \( k \) sensor-derived measurements in the sensing data component. The observation from the \( i\)th sensor is denoted as \( \mathbf{x}_i = [x_{i,1}, \dots, x_{i,t}]^\top \), where $t$ is the sampling duration. Our proposed framework, AttenGluco, leverages CGM data ($\mathbf{x}_{\text{g}}$) and activity data such as walking steps ($\mathbf{x}_{\text{ws}}$) and walking time intervals ($\mathbf{x}_{\text{wi}}$), which represent the duration between consecutive walking events. The multi-step forecasting output is expressed as $ \mathbf{\hat{x}}_g = [x_{g,t+1}, \dots, x_{g,t+m}]^\top $, where $m$ represents the number of predicted time steps, commonly referred to as the prediction horizon (PH). Mathematically, the forecasting task can be formulated as $\mathbf{\hat{x}}_g = f(\mathbf{X; \Theta})$, where \( f \) represents the forecasting model, parameterized by \( \Theta \), which is learned during the training process.

\subsection{AttenGluco}

AttenGluco is composed of two primary stages. The first stage, data preparation, focuses on collecting and processing physiological and behavioral data to serve as input for the forecasting model. This stage also includes data interpolation to handle missing values and normalization for consistency. During this phase, BGLs are recorded using a CGM device, while additional behavioral metrics, such as physical activity, are gathered from wearable sensors such as smartwatches, as depicted in Fig. \ref{fig:proposed_method}. The second stage is the multimodal forecasting model, which utilizes these preprocessed inputs for blood glucose prediction.

\begin{figure}[t]
    \centering
    \includegraphics[width=0.5\textwidth]{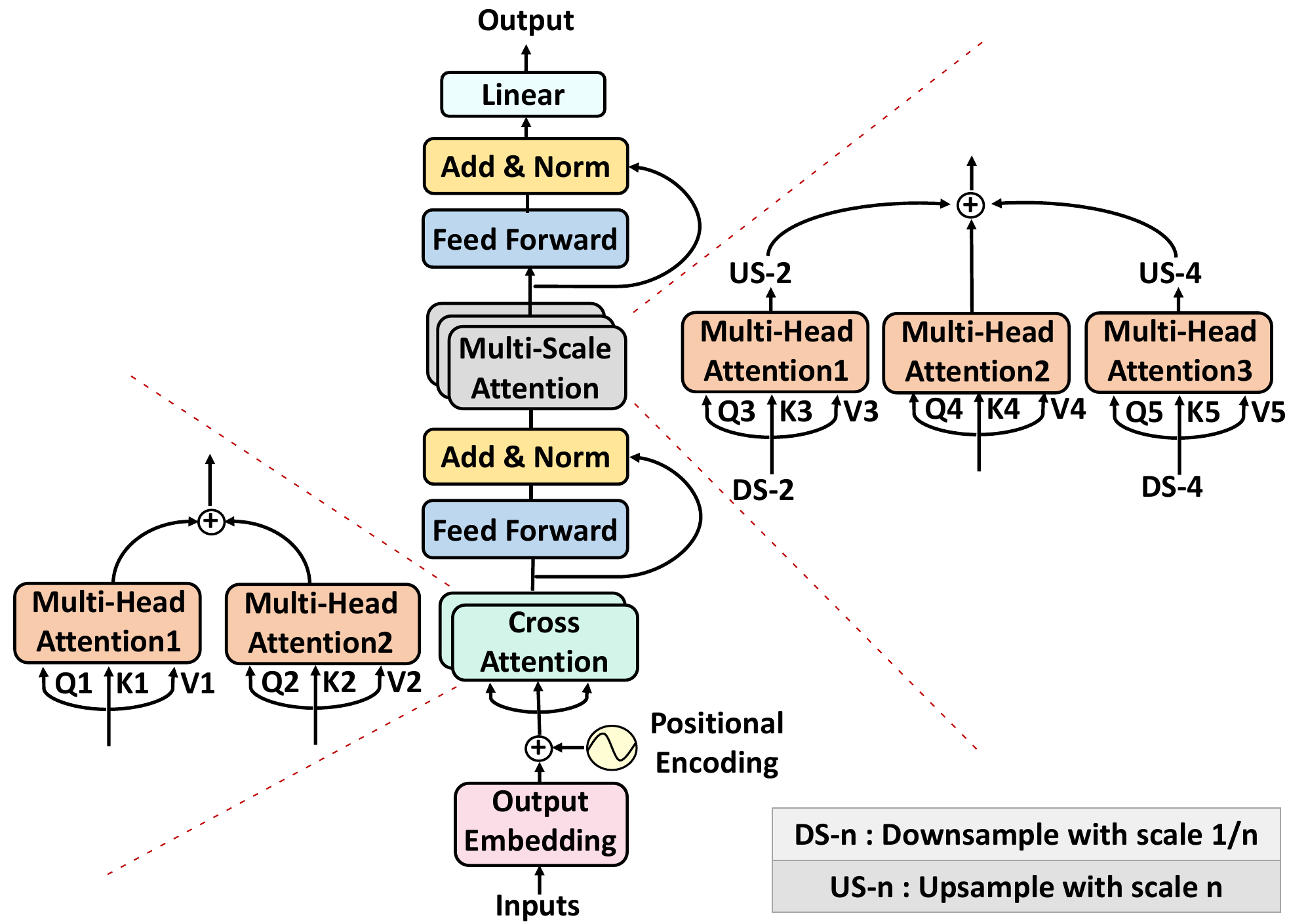}
    \caption{\small AttenGluco model architecture consists of cross-attention and multi-scale attention to forecast BGL.}
    \label{fig:transformer_method}
\end{figure}

The forecasting model is developed based on the Transformer architecture. This architecture leverages an attention mechanism to extract time-dependent patterns from fused irregular time-series data while also capturing long-term dependencies. This approach enables the model to effectively process complex temporal relationships. The structure of our proposed Transformer-based forecasting model is shown in Fig.~\ref{fig:transformer_method}.

The standard transformer architecture is typically made up of an encoder-decoder for data reconstruction. However, we modified this design for our forecasting model and framed it as a supervised learning task. Specifically, we eliminated the decoder and only utilized the encoder for data representation learning. Our customized transformer architecture incorporates two attention mechanisms: cross-attention and multi-scale attention. The cross-attention mechanism integrates various time series data with variant sample rates, while the multi-scale attention captures temporal dependencies within the signals to reduce the effect of random noise~\cite{shabani2022scaleformer}. By incorporating these attention mechanisms, our Transformer-based approach enhances the accuracy of BGL forecasting.

Our Transformer architecture consists of embedding and positional encoding layers, followed by cross-attention, feed-forward, Add \& Norm layers, and a multi-scale attention block. The input variables \( \mathbf{x}_\text{g} \), \( \mathbf{x}_\text{ws} \), and \( \mathbf{x}_\text{wi} \) are initially processed through an embedding layer \( f_{\text{embed}}(\cdot) \), then passed through a positional encoding function \( f_{\text{pos}}(\cdot) \), producing the transformed representations \( \mathbf{X}_\text{G} \), \( \mathbf{X}_\text{WS} \), and \( \mathbf{X}_\text{WI} \), respectively. Each resulting matrix resides in \( \mathbb{R}^{t \times d_{\text{model}}} \), where \( t \) represents the sampling duration and \( d_{\text{model}} \) is a hyperparameter. The multi-head attention mechanism in Transformer architectures~\cite{vaswani2017attention} functions by scaling values \( (\mathbf{V} \in \mathbb{R}^{t \times d_\text{model}}) \) based on the relationships between keys \( (\mathbf{K} \in \mathbb{R}^{t \times d_{\text{model}}}) \) and queries \( (\mathbf{Q} \in \mathbb{R}^{t \times d_{\text{model}}}) \). The mathematical formulation of the attention mechanism is presented in Eq.~\ref{eq:attention}.

\begin{equation}
\text{Attention}(\mathbf{Q},\mathbf{K},\mathbf{V})= \text{Softmax} \left(\frac{\mathbf{Q} \mathbf{K}^T}{\sqrt{d_{\text{model}}}} \right)\mathbf{V}
\label{eq:attention}
\end{equation}
We designed a two-branch cross-attention layer, where both branches receive \( \mathbf{X}_\text{G} \) as the query. In one branch, the keys and values correspond to \( \mathbf{X}_{\text{WS}} \), while in the other, they correspond to \( \mathbf{X}_{\text{WI}} \). The cross-attention (CA) of the first branch is computed using Eqs.~\ref{eq:crossatt} and~\ref{eq:crossatt2}.

\begin{equation}
        \text{CA}\left(\mathbf{X_\text{G}}, \mathbf{X_{\text{WS}}}, \mathbf{X_{\text{WS}}}\right) \\
         =[\mathbf{H}_1, \dots, \mathbf{H}_{m_H}] \mathbf{W}_H^{\text{CA}}
    \label{eq:crossatt}
\end{equation}
\begin{equation}
    \mathbf{H}_h = \text{Attention}(\mathbf{\mathbf{X_\text{G}}} \mathbf{W}_{\mathbf{Q}}^{\text{CA}}, \mathbf{X}_\text{WS} \mathbf{W}_{\mathbf{K}}^{\text{CA}}, \mathbf{X}_\text{WS} \mathbf{W}_{\mathbf{V}}^{\text{CA}})
    \label{eq:crossatt2}
\end{equation}

Where $\mathbf{W}_{\mathbf{Q}}^{\text{CA}}$, $\mathbf{W}_{\mathbf{K}}^{\text{CA}}$, and  $\mathbf{W}_{\mathbf{V}}^{\text{CA}}$ are weight matrices specific to the attention head and belong to $\mathbb{R} ^{d_{\text{model}}\times d_{\text{model}}}$. Moreover, $\mathbf{W}_H^{\text{CA}} \in \mathbb{R}^{(m_H \cdot d_{\text{model}}) \times d_{\text{model}}}$ is the final weight matrix that projects the concatenated attention head outputs into the original model dimension.The attention mechanism for the second branch follows the same computation, with the $\mathbf{X}_\text{WI}$ as both the key and the query.

Then, the attention outputs from both branches are combined to incorporate cross-attention information. The resulting data is passed through a linear feedforward network followed by an Add \& Norm module. The processed output, \( \mathbf{X}_{\text{CA}} \), is then fed into a multi-scale attention mechanism comprising three multi-head attention branches, each designed for different downsampling (DS) rates. These branches apply downsampling factors of 1, 2, and 4, where a factor of 1 indicates no downsampling, as illustrated in Fig.~\ref{fig:transformer_method}.

For the first branch, the multi-scale attention mechanism (MA) on $\mathbf{X}_{\text{CA}}$ is computed by using Eqs.~\ref{eq:multiatt} and~\ref{eq:multiatt2}.
\begin{equation}
        \text{MA}(\mathbf{X}_\text{CA}, \mathbf{X}_\text{CA}, \mathbf{X}_\text{CA}) =
        [\mathbf{H}_1, \dots, \mathbf{H}_{m_H}] \mathbf{W}_H^{\text{MA}}
    \label{eq:multiatt}
\end{equation}
\begin{equation}
    \mathbf{H}_h = \text{Attention}(\mathbf{X}_\text{CA} \mathbf{W}_{\mathbf{Q}}^{\text{MA}},\mathbf{X}_\text{CA} \mathbf{W}_{\mathbf{K}}^{\text{MA}}, \mathbf{X}_\text{CA} \mathbf{W}_{\mathbf{V}}^{\text{MA}})
    \label{eq:multiatt2}
\end{equation}

Each attention branch utilizes query, key, and value weight matrices, \( \mathbf{W}_{\mathbf{Q}}^{\text{MA}} \), \( \mathbf{W}_{\mathbf{K}}^{\text{MA}} \), and \( \mathbf{W}_{\mathbf{V}}^{\text{MA}} \), all belonging to \( \mathbb{R} ^{d_{\text{model}}\times d_{\text{model}}} \). The outputs from all attention heads are concatenated and projected back into the original model dimension using the final weight matrix \( \mathbf{W}_H^{\text{MA}} \in \mathbb{R}^{(m_H \cdot d_{\text{model}}) \times d_{\text{model}}} \). The remaining two branches follow the same computational process but operate on downsampled input data. This approach improves the model’s capability to capture both fine-grained details and long-term temporal dependencies within the input signals.

The outputs from the three multi-scale attention branches are summed and passed through a feed forward network, an Add \& Norm block, and a fully connected layer. This final configuration generates \( m \) predicted CGM values. Each prediction corresponds to a measurement taken every 5 minutes, meaning that \( m \) samples collectively provide forecasts for \( m \times \) 5 minutes into the future. In summary, Algorithm~\ref{alg:TransformerBGL} describes the data processing pipeline in AttenGluco.

\begin{algorithm}
\small
\caption{AttenGluco model}
\label{alg:TransformerBGL}
\textbf{Input:} Preprocessed and normalized data, including CGM signal ($\mathbf{x}_{\text{g}}$), walking steps ($\mathbf{x}_{\text{ws}}$), and walking time intervals ($\mathbf{x}_{\text{wi}}$), Cross-attention block (CA), Multi-scale atention block (MA), Embedding function ($f_{\text{embed}}$), Positional encoding function ($f_{\text{pos}}$), Two Add \& Norm block ($f_{\text{AN}}^{(1)}$,$f_{\text{AN}}^{(2)}$), Two Feedforward model ($f_\text{FF}^{(1)}$,$f_\text{FF}^{(2)}$), Linear model ($f_\text{lin}$) \\
\textbf{Output:} Predicted BGL $\mathbf{\hat{x}}_g$\\
\begin{algorithmic}[1]
\State \textbf{Begin}

\vspace{0.5em}
\State \hspace{0.2cm} $[\mathbf{X}_\text{G},\mathbf{X}_\text{WS}, \mathbf{X}_\text{WI}]\gets f_{\text{pos}}\left(f_{\text{embed}}\left(\left[\mathbf{x}_\text{g}, \mathbf{x}_\text{ws}, \mathbf{x}_\text{wi}\right]\right)\right)$
\vspace{0.5em}
\State \hspace{0.2cm} $\mathbf{X}_{\text{CA1}}\gets$ CA($\mathbf{X}_{\text{G}}, \mathbf{X}_{\text{WS}}, \mathbf{X}_{\text{WS}}$)
\vspace{0,5em}
\State \hspace{0.2cm} $\mathbf{X}_{\text{CA2}}\gets$ CA($\mathbf{X}_{\text{G}}, \mathbf{X}_{\text{WI}}, \mathbf{X}_{\text{WI}}$)
\vspace{0.5em}
\State \hspace{0.2cm} $\mathbf{X}_\text{CA}\gets f_{\text{AN}}^{(1)}\left(f_{\text{FF}}^{(1)}\left(\mathbf{X}_\text{CA1}+\mathbf{X}_\text{CA2}\right)\right)$
\vspace{0.5em}
\State \hspace{0.2cm}  $\mathbf{X}^{(2)}_{\text{CA}}, X^{(4)}_{\text{CA}} \gets \text{Downsample}(\mathbf{X}_{\text{CA}}, 2), \text{Downsample}(\mathbf{X}_{\text{CA}}, 4)$
\vspace{0.5em}
\State \hspace{0.2cm} $\mathbf{X}_{\text{MA1}}\gets$ MA$(\mathbf{X}_{\text{CA}}, \mathbf{X}_{\text{CA}}, \mathbf{X}_{\text{CA}})$
\vspace{0.5em}
\State \hspace{0.2cm} $\mathbf{X}_{\text{MA2}}\gets$ Upsample(MA$(\mathbf{X}^{(2)}_{\text{CA}}, \mathbf{X}^{(2)}_{\text{CA}}, \mathbf{X}^{(2)}_{\text{CA}})$, 2)
\vspace{0.5em}
\State \hspace{0.2cm} $\mathbf{X}_{\text{MA3}}\gets$ Upsample(MA$(\mathbf{X}^{(4)}_{\text{CA}}, \mathbf{X}^{(4)}_{\text{CA}}, \mathbf{X}^{(4)}_{\text{CA}})$, 4)
\vspace{0.5em}
\State \hspace{0.2cm} $\mathbf{X}_\text{MA}\gets f_{\text{AN}}^{(2)}\left(f_{\text{FF}}^{(2)}\left(\mathbf{X}_\text{MA1}+\mathbf{X}_\text{MA2}+\mathbf{X}_\text{MA3}\right)\right)$
\vspace{0.5em}
\State \hspace{0.2cm} $\hat{\mathbf{x}}_g \gets f_\text{lin}(\mathbf{X}_\text{MA})$
\vspace{0.5em}
\State \hspace{0.2cm} \textbf{return} $\hat{\mathbf{x}}_g$
\State \textbf{End}
\end{algorithmic}
\vspace{-0.8mm}
\end{algorithm}

\section{Results \& Discussion}
In this section, we first introduce the AI-READI dataset used to train AttenGluco. We then compare its performance against a baseline model consisting if a 1D-CNN and LSTM for blood glucose forecasting to highlight the significance of our model for providing accurate forecasting. The baseline model, a multimodal LSTM, is commonly employed in state-of-the-art blood glucose prediction. The comparison is conducted using error metrics, including Root Mean Square Error (RMSE) and Mean Absolute Error (MAE), as well as correlation analysis. We investigate various training and testing scenarios to comprehensively evaluate the performance of AttenGluco.

\subsection{Dataset Description}
\label{section:dataset}
The dataset used in this study is the publicly available AI-READI Flagship Dataset. This dataset is designed to advance AI and machine learning research on Type 2 Diabetes Mellitus (T2DM). Collected from 1,067 participants across three U.S. sites. It includes individuals with and without T2DM, balanced across sex, race, and diabetes severity. The dataset consists of four categories: healthy individuals, individuals with prediabetes, individuals with T2DM on oral medication, and individuals with T2DM on insulin.

A key feature of the dataset is its multi-modal structure, where participants were monitored over ten days using a Dexcom G6 CGM for real-time blood glucose, a Garmin Vivosmart 5 for physical activity and heart rate variability, and a LeeLab Anura sensor for environmental factors such as air quality and temperature. The dataset also includes survey data, clinical assessments, and retinal imaging. Daily step counts are recorded via an accelerometer, with occasional gaps due to device recharging. The heart rate sensor also computed a stress index (0-100) based on heart rate variability.

For this study, CGM data and walking activity (steps and intervals) are extracted as key features. After filtering out subjects with missing data, 896 participants are included in the final analysis, distributed as follows: 323 healthy individuals, 207 pre T2DM, 258 with T2DM on oral medication, and 108 with T2DM on insulin.

\subsection{Experimental Setup}
The baseline model follows a 1D-CNN architecture coupled with an LSTM. The 1D-CNN consists of two convolutional layers with 64 and 128 filters, each using a kernel size of 3. This is followed by a two-layer LSTM with 128 and 64 output features. The LSTM output is then passed through an MLP composed of three fully connected layers. 

Both AttenGluco and the baseline model receive a sliding window of historical data covering 6.66 hours (400 minutes) as input. Training is conducted for 300 epochs with a learning rate of 0.001, optimizing the Mean Squared Error (MSE) using the Adam optimizer. Forecasting performance is assessed across three prediction horizons (PHs): 5 minutes, 30 minutes, and 60 minutes. To ensure consistency, each model undergoes five independent training runs. Model performance is assessed across all subjects, with comparisons based on RMSE~\cite{arefeen2023glysim}, MAE~\cite{arefeen2023glysim}, and Correlation~\cite{zhang2023joint}.

As mentioned in section~\ref{section:dataset}, the AI-READI dataset categorizes subjects into four cohorts (healthy, pre-T2DM, oral, and insulin) based on diabetes severity. To evaluate AttenGluco’s performance across these cohorts, we conducted three distinct experiments under different scenarios (subject training, cohort-wise fine-tuning, and forgetting analysis) and compared the results with the baseline model. The details of each scenario will be discussed in the following sections.

\subsubsection{Isolated Subject Training}
In this scenario, the CGM and activity data of AI-READI participants are first grouped according to their respective cohorts. The proposed model is then applied to each subject individually, with 85\% of their data used for training and the remaining 15\% reserved for testing. After evaluating one subject, the model is reinitialized before being trained and tested on the next. Table~\ref{tab:performance Person} presents the average error metrics for AI-READI participants across each cohort separately.
\vspace{-2mm}
\begin{table}[h]
\captionsetup{font=small} 
\caption{Comparison of baseline and AttenGluco performance across different cohorts in the isolated subject scenario. The best results are highlighted in bold.}
    \centering
    \renewcommand{\arraystretch}{1} 
    \setlength{\tabcolsep}{2pt} 
    \begin{adjustbox}{max width=\columnwidth}
    \begin{tabular}{|c|cc|cc|cc|}
        \hline
        \multirow{2}{*}{\textbf{Cohort}} & \multicolumn{2}{c|}{\textbf{RMSE}} & \multicolumn{2}{c|}{\textbf{MAE}} & \multicolumn{2}{c|}{\textbf{Correlation}} \\
        & Baseline & AttenGluco & Baseline & AttenGluco & Baseline & AttenGluco \\
        \hline
        Healthy     & 18.04 & \textbf{16.05} & 13.02  & \textbf{11.12} & 0.38 & \textbf{0.49} \\
        Pre-T2DM & 19.95 & \textbf{18.27} & 15.12  & \textbf{13.65} & 0.49 & \textbf{0.57} \\
        Oral        & 25.01  & \textbf{22.56} & 17.9   & \textbf{15.74} & 0.55 & \textbf{0.64} \\
        Insulin      & 29.9  & \textbf{27.18} & 22.28  & \textbf{19.93} & 0.59 & \textbf{0.67} \\
        \hline
    \end{tabular}
    \label{tab:performance Person}
    \end{adjustbox}
\end{table}

Table~\ref{tab:performance Person} depicts that our proposed method surpassed the baseline model in all performance metrics. For instance, compared to the baseline model, AttenGluco improves the RMSE metric by 11.03\%, 8.42\%, 9.79\%, and 9.09\% for the healthy, pre-T2DM, oral, and insulin cohorts, respectively.
\subsubsection{Cohort-Wise Fine-Tuning}
In the  cohort-wise fine-tuning scenario, the model is trained progressively within each participant category, unlike the isolated subject scenario where it is reset for each subject. Here, the model is first trained on one subject and then fine-tuned sequentially across the other subjects in the same category, with each subject serving as both training and testing data. This process continues until all subjects in a category have been used. Once a category is completed, the model is reinitialized before moving on to the next cohort. The average performance metrics for each category are presented in Table~\ref{tab:performance}. This approach enables the model to gradually adapt to variations within each cohort; therefore, it achieves better performance than the previous scenario.

\begin{table}[h]
    \centering
    \renewcommand{\arraystretch}{1.2} 
    \setlength{\tabcolsep}{6pt} 
    \captionsetup{justification=centering, font=small} 
    \caption{Performance comparison between the baseline model and AttenGluco across different cohorts in the cohort-wise fine-tuning scenario. The best scores are highlighted in bold.}
    \label{tab:performance}
    
    \begin{adjustbox}{max width=\columnwidth}
    \begin{tabular}{|l|cc|cc|cc|}
        \hline
        \multirow{2}{*}{\textbf{Cohort}} & \multicolumn{2}{c|}{\textbf{RMSE}} & \multicolumn{2}{c|}{\textbf{MAE}} & \multicolumn{2}{c|}{\textbf{Correlation}} \\
        & \textbf{Baseline} & \textbf{AttenGluco} & \textbf{Baseline} & \textbf{AttenGluco} & \textbf{Baseline} & \textbf{AttenGluco} \\
        \hline
        Healthy      & 17.79  & \textbf{15.45}   & 12.79  & \textbf{10.96}  & 0.44  & \textbf{0.53} \\
        Pre-T2DM & 19.77  & \textbf{17.47}   & 14.41  & \textbf{12.46}  & 0.51  & \textbf{0.6} \\
        Oral         & 23.37  & \textbf{20.45}   & 16.93  & \textbf{14.71}  & 0.57  & \textbf{0.67} \\
        Insulin      & 28.22  & \textbf{25.04}   & 20.51  & \textbf{18.03}  & 0.68  & \textbf{0.75} \\

        \hline
    \end{tabular}
    \end{adjustbox}
\end{table}
Referring to Table~\ref{tab:performance}, we conclude that AttenGluco outperforms the baseline model across all performance metrics. It improves RMSE by 13.15\%, 11.63\%, 12.49\%, and 11.27\% for the healthy, pre-T2DM, oral, and insulin cohorts, respectively.

\begin{figure}
    \centering
    \includegraphics[width=\linewidth]{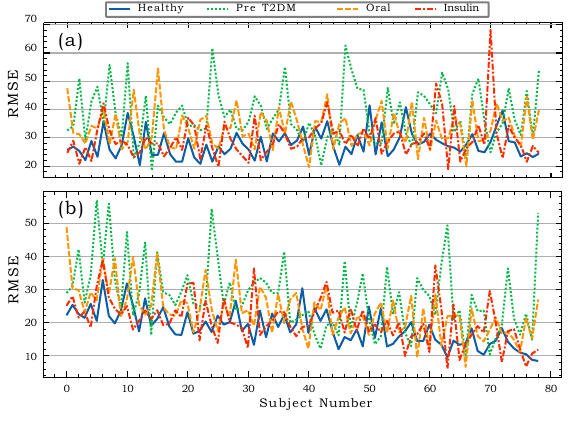}
    \caption{\small The RMSE of each subject in (a) the baseline model and (b) AttenGluco under the cohort-wise scenario. As observed, AttenGluco's RMSE decreases significantly within each cohort as more subjects are incorporated into the fine-tuning process.}
    \label{fig:trendline}
\end{figure}

Fig.~\ref{fig:trendline} demonstrates that as more subjects are added into each cohort, the model's performance progressively improves in this scenario. This results in lower errors for newer subjects when used for testing. Notably, the reduction in test error is more significant in AttenGluco, indicating that its performance could further improve with a larger training dataset. For improved clarity and better visibility, we illustrate only 80 subjects of each cohort in Fig.~\ref{fig:trendline} while maintaining the overall distribution and trends of the complete dataset.

Moreover, we evaluate the AttenGluco's forecasting RMSE at different PH values of $5$, $30$, and $60$ minutes. Since CGM data is recorded at 5-minute intervals, a PH of 5 minutes corresponds to \( m = 1 \) sample, a PH of 30 minutes corresponds to \( m = 6 \) samples, and a PH of 60 minutes corresponds to \( m = 12 \) samples. Table~\ref{tab:rmse_comparison} presents a comparison of AttenGluco and the baseline model across different PHs. As shown, increasing the PH leads to a higher RMSE for both models. However, while the baseline model experiences a significant drop in performance, AttenGluco maintains a relatively stable RMSE. This demonstrates AttenGluco's robustness in long-term forecasting.
\begin{table}[h]
\vspace{-1.5mm}
    \centering
    \renewcommand{\arraystretch}{1.2} 
    \setlength{\tabcolsep}{8pt} 
    \caption{\small RMSE Comparison of Baseline and AttenGluco Models Across Different PHs}
    \label{tab:rmse_comparison}
     \begin{adjustbox}{max width=\columnwidth}
    \begin{tabular}{|l|ccc|ccc|}
        \hline
        \multirow{2}{*}{\textbf{Cohort}} & \multicolumn{3}{c|}{\textbf{Baseline RMSE}} & \multicolumn{3}{c|}{\textbf{AttenGluco RMSE}} \\
        & \textbf{5 min} & \textbf{30 min} & \textbf{60 min} & \textbf{5 min} & \textbf{30 min} & \textbf{60 min} \\
        \hline
        Healthy        & 7.35  & 14.37  & 17.79  & 7.63  & 12.38  & 15.45 \\
        Pre-T2DM   & 7.94  & 15.43  & 19.77  & 8.70  & 13.50  & 17.47 \\
        Oral           & 9.15  & 17.73  & 23.37  & 9.33  & 15.21  & 20.45 \\
        Insulin        & 12.11 & 21.00  & 28.22  & 11.94 & 18.55  & 25.04 \\
        \hline
    \end{tabular}
    \end{adjustbox}
\end{table}

\subsubsection{Continual Learning and Forgetting Analysis}
Even though transferring the model to and fine-tuning it on new subjects enhances the model's performance on new data, it simultaneously leads to the loss of previously learned knowledge. This phenomenon, known as catastrophic forgetting~\cite{kirkpatrick2017overcoming}, is a well-known issue that happens with model retraining. The problem becomes more pronounced when there is a significant distribution shift between the old and new data, which causes the model to prioritize recent patterns while disregarding past ones.

\begin{figure}
\vspace{-2mm}
    \centering
    \includegraphics[width=\linewidth]{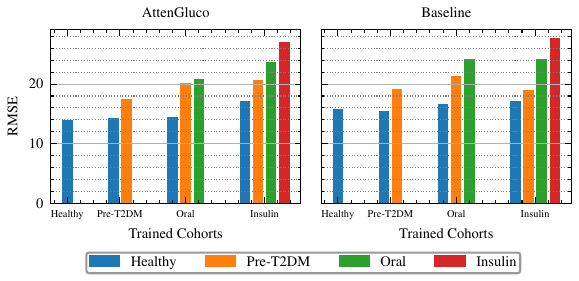}
    \caption{Fine-tuning the model on new cohorts leads to the loss of knowledge from previous ones.}
    \label{fig:cl_results}
    \vspace{-2mm}
\end{figure}

We hypothesize that a distribution shift exists among the four cohorts, which potentially causes the model to forget previously learned information as new cohorts are introduced. To measure forgetting in both models, we evaluate their performance on prior cohorts after completing training on new ones. In this scenario, the model continues training on all subjects of the cohorts without reinitialization. The results are presented in Fig.~\ref{fig:cl_results}, where the x-axis represents the training cohorts, and each grouped bar chart illustrates the model’s performance after training on the respective cohort. As shown, the introduction of new cohorts degrades both model’s retention of previous knowledge.
\section{Conclusion}
In this study, we proposed AttenGluco, a multimodal Transformer-based framework for long-term blood glucose forecasting using CGM and activity data. By integrating cross-attention and multi-scale attention, our model effectively fuses heterogeneous time-series data and captures long-term dependencies. Our evaluation on the AI-READI dataset demonstrated that AttenGluco outperforms baseline models under different test and train scenarios across various subject cohorts. AttenGluco improved RMSE by about 10\% in the isolated subject training scenario. In the cohort-wise fine-tuning scenario, RMSE improvements are even more pronounced, with reductions of about 12\%. Additionally, AttenGluco achieved higher correlation scores across all groups, further validating its enhanced predictive capability. Our analysis of forecasting accuracy at different prediction horizons (5, 30, and 60 minutes) shows that AttenGluco consistently outperformed the baseline model, with the most notable gains observed at longer horizons, where it reduced RMSE by up to 3.18 compared to the baseline. Furthermore, our forgetting analysis revealed that AttenGluco maintains lower error rates when fine-tuned on new cohorts. By improving long-term blood glucose forecasting, AttenGluco has the potential to advance precision medicine for diabetes care, enabling more proactive and individualized interventions to maintain optimal glucose levels.


\bibliographystyle{IEEEtran}
\bibliography{refs}

\end{document}